\documentclass[11pt]{article}
\usepackage[final]{acl}
\usepackage{times}
\usepackage{latexsym}
\usepackage{tabularx}
\usepackage[T1]{fontenc}
\usepackage[utf8]{inputenc}
\usepackage{microtype}
\usepackage{inconsolata}
\usepackage{graphicx}
\usepackage{enumitem} 
\graphicspath{ {./images/} }
\usepackage{float}
\usepackage{arydshln} \usepackage{stfloats} 
\usepackage{times}

\title{DOLFIN - \underline{Do}cument-\underline{L}evel \underline{Fin}ancial test set for Machine Translation}

\author{Mariam Nakhlé$^{1,2}$, Marco Dinarelli$^1$, \textbf{Raheel Qader}$^2$, \\ 
\textbf{Emmanuelle Esperança-Rodier}$^1$,  \textbf{Hervé Blanchon}$^1$ \\ 
  (1) Univ. Grenoble Alpes, CNRS, Grenoble INP \footnote{Institute of Engineering Univ. Grenoble Alpes}, LIG, 38000 Grenoble, France \\
  (2) Lingua Custodia, 75008, Paris, France \\
  \texttt{contact: mariam.nakhle@univ-grenoble-alpes.fr} }

\begin{document}
\maketitle
\begin{abstract}

Despite the strong research interest in document-level Machine Translation (MT), the test sets dedicated to this task are still scarce. The existing test sets mainly cover topics from the general domain and fall short on specialised domains, such as legal and financial. Also, in spite of their document-level aspect, they still follow a sentence-level logic that does not allow for including certain linguistic phenomena such as information reorganisation.
In this work, we aim to fill this gap by proposing a novel test set: DOLFIN. The dataset is built from specialised financial documents, and it makes a step towards true document-level MT by abandoning the paradigm of perfectly aligned sentences, presenting data in units of sections rather than sentences.
The test set consists of an average of 1950 aligned sections for five language pairs. We present a detailed data collection pipeline that can serve as inspiration for aligning new document-level datasets.
We demonstrate the usefulness and quality of this test set by evaluating a number of models. 
Our results show that the test set is able to discriminate between context-sensitive and context-agnostic models and shows the weaknesses when models fail to accurately translate financial texts.
The test set is made public for the community\footnote{\url{https://huggingface.co/datasets/LinguaCustodia/dolfin}}.
\end{abstract}

\section{Introduction}

\begin{table}
\centering
\begin{tabular}{llll}
\hline
\textbf{Language} & \textbf{\# Seg.}   & \textbf{\# Avg sent.}  \\
\hline
En-Fr &    2081  & 12.1  \\
En-De &    2026  & 11.4  \\
Fr-Es &    1983  & 15.2  \\
En-Es &    1932  & 14.8  \\
En-It &    1737  & 10.8  \\
\hline
Average &    1951  & 12.8  \\
\hline
\end{tabular}
\caption{\label{table:stats_langs}Number of segments and sentences per language pair. \# Seg. is the number of segments per language in the dataset and \# Avg sent. is the average number of sentences in a segment.
}
\end{table}

Document-level translation has received a lot of attention in the Machine Translation (MT) community in the past years \cite{toral2018attaining, laubli2020set}. However, document-level testing data is still scarce, covering mainly the news domain and only a few specialised domains, which hinders progress \cite{anastasopoulos-etal-2020-tico, goyal2022flores, federmann-etal-2022-ntrex}. 

In this work, we focus on the financial domain which is characterised by a highly specialised language and a high demand for fast and reliable translations. The translations of some of the financial documents are required by law, which creates a significant demand for translation services with a high degree of in-domain expertise. A salient example of this is the French word \textit{couverture} that translates as \textit{blanket} in a general domain text, but as \textit{hedge} in a financial text.
This makes out of financial translation an interesting use-case for MT. In order to facilitate a robust context-sensitive evaluation within this domain, we propose \emph{DOLFIN}, a new document-level test set that focuses on specialised language from the financial domain. Beside the novelty of the domain, this test set also employs a novel approach to defining translation units: rather than considering aligned sentences, it considers aligned sections, thereby enabling higher-level information reorganisation, such as sentence reordering within the section. A further innovation is that this test set was built with a focus on context-sensitive phenomena, which makes it an approach in-between a regular test set and a test-suite.

As was shown in \citet{toral2018attaining} and \citet{laubli2020set}, sentence-level MT suffers from errors that are only identifiable when taking the context into consideration. This is true for the financial domain as well. In specialised domains, terminology plays an important role and the terms used must be consistent throughout the document, therefore one can only assess the correctness of a term when having access to its previous and following translations. This is especially true for legal financial documents, where the beginning of a document contains an explicit definition of terms used for the mentioned entities that must be respected throughout the document. Another example would be the consistency of numerical formats. Financial documents contain many tables stating monetary amounts and using different norms of formatting within the same table would be unacceptable (for example \textit{5 million USD}, \textit{5M USD} or \textit{\$5M}). These phenomena are why a document-level evaluation is especially relevant for financial MT.

We use the Fundinfo website\footnote{\url{https://www.fundinfo.com}} as a source of data because it offers a vast choice of parallel financial documents in numerous European languages. The \emph{Terms of Use} of this website allow us to crawl the documents where a download-link is provided\footnote{\url{https://api.fundinfo.com/static/legal/1/en}, last consulted on 09/09/2024.}. The documents on Fundinfo  gather all the official translations for the given regions and languages.

In this work we are contributing to the goal of a robust document-level MT evaluation in the specialised financial domain by collecting a test set covering 5 language pairs with 1951 segments on average per language pair, as shown in Table \ref{table:stats_langs}. The contributions of this work can be summarised as following: \\
(1) we create and publicly release DOLFIN, a novel test set for document-level financial MT, \\
(2) we describe a detailed pipeline to extract and align sections from long documents, \\
(3) we propose an approach to measure the impact of longer context on translation quality by comparing the scores obtained in a sentence-level and in a document-level fashion, \\ 
(4) we present an experiment showing the usefulness of this test set and analyse the capabilities of LLMs to deal with longer contexts and specialised financial language.

\section{Related work}

\label{sec:related_work}

\textbf{Document-level test sets}

Most resources in MT research are sentence-level, but there is a number of test sets that include not only individual sentences but also longer texts. These are usually made without trying to target any context-sensitive phenomenon and typically the evaluation is performed at sentence-level, i.e., by feeding the sentences separately to the evaluation metrics \cite{anastasopoulos-etal-2020-tico, goyal2022flores, federmann-etal-2022-ntrex,deutsch-etal-2023-training}.
These works usually take the form of a source file, target file and a file with document ids. The order of lines creates a mapping between the source and target sentences as well as the document ids, therefore all three files must contain the same number of lines.

Our test set diverges by taking a novel approach and abandoning the idea of a perfect sentence-to-sentence alignment. We consider that it constrains the translation process and strips the texts of high-level phenomena such as information reorganisation, sentence splitting and merging. 
Section \ref{sec:imperf-alignment} details the reasons behind this choice.

There are also a number of document-level training datasets \cite{koehn-2005-europarl, cettolo2012wit3, tiedemann-2012-parallel, lison-tiedemann-2016-opensubtitles2016} and recently there have been efforts to recover the contexts of large sentence-level datasets \cite{wicks2024recovering}. Since these are for training purposes, they are out of the scope of this work that focuses on test data.

As for datasets specialised on financial language, these are very scarce; we can mention the European Central Bank corpus that is included in OPUS \cite{tiedemann_opus} and a diachronic corpus based on a banking magazine \cite{volk_banking}. However, these are resources are the sentence level.

\textbf{Targeted context-sensitive evaluation}

An important line of work in the document-level MT evaluation are test-suites. These are manually crafted test sets that target specific context-sensitive phenomena and the evaluation focuses on errors when translating them. There are two approaches: 1) evaluation based on references and 2) evaluation based on contrastive pairs.

The first type is inspired by the traditional approach of comparing the generated translation to a reference \cite{guillou-etal-2016-findings, wong-kit-2012-extending, hardmeier-federico-2010-modelling}.
The use of references has the disadvantage that it only measures the agreement between the reference and hypothesis and not the internal agreement across generated sentences, i.e., whether the translation respects agreement with its previous and following sentences.

To overcome this, the contrastive pairs were proposed. For every source sentence, multiple context sentences are provided, along with a correct translation, as well as an incorrect one (or multiple incorrect ones) \cite{bawden-etal-2018-evaluating, muller-etal-2018-large, voita-etal-2019-good}. The evaluation measures if the model gives a higher probability to the correct translation without actually requiring the model to generate it. The disadvantage of these test-suites is the need to access models' probabilities, which is impossible for most commercial systems, and another flaw is that even if the model gives a higher probability to the correct answer, it doesn't guarantee that the generated translation would be correct. 
To build new targeted test-suites more easily, automatic means to identify sentences that contain context-sensitive phenomena were proposed  \cite{fernandes-etal-2023-translation, wicks-post-2023-identifying}.

The development of new test-suites usually requires a high level of linguistic expertise, knowledge about the studied context-sensitive phenomena and/or tools designed for the task. Therefore, test-suites are only available for some languages and well studied phenomena, which makes the evaluation limited. This makes the test-suites difficult to use as a universal method for document-level MT evaluation. This issue is usually bridged by combining a test-suite evaluation with a general quality test set evaluation. However, a more ideal approach would be being able to capture the models' capabilities to deal with context, as well as its overall translation quality. 
In the following, we present the DOLFIN test set that aims to evaluate both of these aspects.

\section{Dataset collection}
\label{sec:dataset_collection}

\subsection{Format and structure}
\label{sec:format_structure}

This test set is built from PDF documents that undergo a processing pipeline with the first step consisting of PDF to text extraction. This step can introduce errors that sometimes affect parts of documents, and rarely even full documents.
This is why the test set is built of sections, rather than full-length documents.
Also, in the financial domain, documents range from a few pages to more than hundreds of pages, it would be unfeasible to include such long documents as single segments.
Since most MT test sets annotated with human judgements are also used to train or meta-evaluate automatic metrics, it is preferable to have a single score per section instead of per document, which would drastically decrease the number of scores (and data available for training and meta-evaluation).

When performing the extraction, we chose markdown (MD) as the format for our segments. It is richer than plain text and conserves more information about formatting, like tables, titles and subsections that are common in the PDF documents.

\subsection{Imperfect sentence to sentence alignment}
\label{sec:imperf-alignment}

As mentioned by~\citet{deutsch-etal-2023-training}, there are practically no test sets with real-life translations without a perfect sentence to sentence alignment. The majority of test sets available in the community, such as those from the annual WMT conference \cite{kocmi-etal-2023-findings}, were crafted for the purpose of evaluating sentence-level MT systems and they maintain a perfect sentence-to-sentence alignment. Even though there isn't a clear consensus in the translators' community on whether the original number of sentences must be kept \cite{baker2018other}, it is clear from the analysis of~\citet{merkel-2001-comparing}, that the number of sentences in the human translations differs from the source. This means that the test sets used in the MT community do not contain some of the more challenging linguistic phenomena such as information and sentence reordering, merging or splitting.
In order to allow these phenomena to happen, we don't enforce a perfect sentence-level alignment, even though it makes the calculation of automatic metrics more challenging. We consider that a truly document-level MT system treats the segment as a whole and won't necessarily generate the same amount of sentences as in the source, therefore we abandon the paradigm of perfectly aligned parallel sentences in the rest of this work.

\subsection{Processing Pipeline}
\label{sec:pipeline}

The only source for this test set is the website Fundinfo, a provider of data about investment funds for different stakeholders of the investment industry. It gathers a large amount of financial documents available in multiple languages and document types. Below, we detail the pipeline designed to process the documents.

\textbf{PDF document alignment.} The PDF files were paired into bilingual documents based on the ISIN code (unique identifier), document type, emission date and language code. The website does not provide any information about which language is the original and which one the translation, for practical reasons we refer to the English document as the source and the other as the target, except for French-Spanish document pairs, where we consider the French as the source.

\textbf{PDF to text extraction.} We used the tool \emph{Apryse}\footnote{\url{https://apryse.com/}, a paid software} to extract text from the PDFs. We chose it because it yielded the best results on our kind of documents in our preliminary tests, compared to the PyMuPDF library. More details about these tests can be found in Appendix \ref{annex:pdf_extractor}.

\textbf{Filtering of problematic documents.} Some documents were flawed and removed as a whole at this stage using the following filters: language identification using XLM-Roberta-base \cite{xlm-roberta}\footnote{papluca/xlm-roberta-base-language-detection} and PDF extraction error identification (texts without any space, texts with a space between every character, gibberish, etc.)

\textbf{Section identification.} In the documents, there is no explicit indication of the beginning and ending of sections. We identified the sections using the markdown structure, such as titles, headings and tables, using regular expressions.

\textbf{Section alignment.} The identified sections were aligned using the LASER contextual sentence embeddings \cite{schwenk-douze-2017-learning} and the \textit{easylaser} library\footnote{\url{https://pypi.org/project/easylaser/}} by calculating the cosine similarity between sentences. Since this model was not trained on long texts, we don't embed full texts but only sentences and align the sections using different heuristics. One of the approaches is to leverage the consecutive titles. If two pairs of consecutive titles have a high laser score, we consider that the section between the two titles is also aligned. When aligning sections that don't start with a title, we use the same logic, but instead of using the titles, we verify if the first and last three sentences are aligned. In both cases we define a threshold of 0.75 for the score, above which we consider that the sentences are aligned. In order to align tables, we only consider documents where the number of tables in the two languages is identical and they have the exact same shape (number of lines and columns). We consider them aligned in the order in which they appear in the document.
This way of aligning the data makes this test set intrinsically document-level, with the main unit being the section rather that the individual sentence. This differs from classical alignment approaches, where only sentences are aligned and those that do not meet a minimum threshold are discarded, thus losing the continuity of the document.

\textbf{Deduplication.} Since the financial documents contain many sections dictated by law, they tend to be repetitive. Therefore we apply harsh deduplication using the minhash algorithm and remove near duplicates. Inspired by~\citet{penedo_dataset_falcon}, we used the library \textit{text-dedup}\footnote{\url{https://github.com/ChenghaoMou/text-dedup}} and this step removed approximately 80\% of the sections. However, it keeps repeated sentences if the segment as a whole is not a duplicate. 

\textbf{Data filtering.} We filtered out noisy segments using the same filters as the ones applied to whole documents. We redo this cleaning because in many cases only some parts of the document are impacted. We also remove segments with a big length difference between source and target language, segments with empty tables, segments where the number of headings in the two languages differ, as well as too short (two sentences or less) and too long segments (more than 4000 characters).

\textbf{Quality Estimation filtering.} 
In order to further filter the samples, we used quality estimation to identify the highest quality segments.
We used the Comet-kiwi model~\cite{rei-etal-2022-Cometkiwi}\footnote{Unbabel/wmt22-Cometkiwi-da}, which is trained to evaluate sentences, but not paragraphs, and has a limited maximum context length. 
In order to score the full segments, we used the SLIDE approach~\cite{raunak-etal-2023-evaluating} with a window size of 3 and stride of 1. This approach would ideally need a perfect sentence to sentence alignment, but we overcome this by using the window size of three, which helps to buffer the differences in number of sentences. Segments shorter than the window size are fed to Comet-kiwi as a whole. We looked for a quality threshold separately for every language pair and selected between 45\

\textbf{Identification of interesting segments.} 
In order to build a test set rich in context-sensitive phenomena to challenge the MT models, we identified such phenomena using two approaches.
First, using the \emph{CTXPRO} tool~\cite{wicks-post-2023-identifying} which identifies gender, formality, and animacy for pronouns, plus verb phrase ellipsis and ambiguous noun inflections.
A limitation of this tool is that it is intended for perfect sentence-to-sentence alignment. We approximated such constraint for segments where the number of sentences in source and target language wasn't the same. We simply concatenated strictly the minimum number of sentences in order to obtain the same number of sentences on both sides.
For the second approach, we prompted a Large Language Model (LLM), namely Llama3-70b~\cite{dubey2024Llama}, to annotate segments that contain elements that need context in order to be correctly translated and evaluated. The full prompt used for the annotation can be found in Annex~\ref{annex:prompts}.

\textbf{Manual filtering.} Based on the annotations from the previous step, we created a selection of candidate segments for every language. As a finishing step, these segments were manually verified and we removed some of the repetitive or less interesting segments.
Not every segment contains a context-sensitive phenomenon, because they don't naturally occur in every text. The segments that don't contain any specific phenomenon are kept because they serve the purpose of assessing the overall translation quality.

\section{The DOLFIN test set}
\label{sec:dolfin}

Table~\ref{table:stats_langs} shows the main statistics of the final test set. There are on average 1950 segments per language pair and each segment contains on average 12 sentences. The number of sentences was calculated using the PySBD segmenter \cite{sadvilkar-neumann-2020-pysbd}. A total of 257706 PDF documents were processed, resulting in a final test set containing 7812274 tokens, with an average of 400 tokens per segment, or 800 tokens source and target text combined. The minimum token length of a segment is 26 (in English) and maximum is 6963 (in Italian).

In finance, there are multiple sub-domains that correspond to different document types. In this test set, 15 document types are represented and Annex~\ref{annex:sub-domains} gives the statistics per document type.

The test set contains the following metadata: 
\texttt{src\_lang}: source language;
\texttt{trg\_lang}: target language;
\texttt{sub\_domain}: sub-domain of finance;
\texttt{date}:  date of publication;
\texttt{Annotation}: annotations of context-sensitive phenomena (obtained by CTXPRO and Llama-3-70b);

Table~\ref{table-examples} shows a few examples of context-sensitive phenomena present in the test set, where the context is needed to correctly disambiguate the sentence to translate, or to maintain a consistent translation along the document.

\begin{table*}[!ht]
\centering
\small
\begin{tabularx}{\textwidth}{XX}
\hline
\underline{\textbf{English text}} \\
The ratings of fixed income securities by credit  rating agencies are a generally accepted  barometer  of credit risk. They are, however, subject to certain  limitations from  an investor’s standpoint. \\
\underline{\textbf{French text}} \\
Les \textbf{notes} accordées aux titres à taux fixe par les  agences de notation sont généralement acceptées  comme le « baromètre » du risque de crédit qu’ils  représentent. Toutefois, \textbf{elles} sont quelque peu \textbf{limitées} du point de vue de l’investisseur. \\
\underline{\textbf{Observation:}} the translation of the anaphoric pronoun ``They'' must respect the  agreement with its antecedent ``ratings'', which is in French a feminine noun ``notes''. \\
\hline 
\underline{\textbf{English text}} \\
| Balance, beginning of period |  | \\
| --- | --- | \\
| - Share class A - USD | 25,000.000 | \\
{| - Share class S - CHF | 85,811.152 |}\\

\underline{\textbf{German text}} \\
| Stand zu Beginn der Berichtsperiode |  \\
| --- | --- | \\
| - Anteilsklasse A - USD | \textbf{25'000.000} | \\
| - Anteilsklasse S - CHF | \textbf{85'811.152} | \\
\underline{\textbf{Observation:}} the formatting of numbers must stay consistent across the document to avoid confusions. In this case, the symbol apostrophe is used as a thousand separator, which is the norm in Swiss German.  \\ 
\hline 
\underline{\textbf{English text}} \\

This product is suitable for investors who ... \\
... have at least a basic knowledge of the financial instruments contained in the fund; \\
... have at least a medium-term investment horizon; \\
... would, in a worst-case scenario, be able to withstand the loss of the entire invested capital. \\
\underline{\textbf{Italian text}} \\
Questo prodotto è adatto agli investitori che ... \\
... \textbf{sono} in possesso di conoscenze almeno elementari in merito agli strumenti finanziari detenuti dal fondo; \\
... \textbf{hanno} almeno un orizzonte d'investimento a medio termine; \\
... \textbf{sono} in grado di sopportare, nel peggiore dei casi, anche la perdita dell'intero capitale investito.\\
\underline{\textbf{Observation:}} the document is written in a condensed writing style, which separates the verbs from their subject, nevertheless the verb-subject agreement must be respected. \\ 
\hline
\underline{\textbf{French text}} \\
À titre indicatif, la performance du Fonds est comparée à l’indice Bloomberg Barclays U.S. Government/Credit Bond Index (Total Return)  (ci-après, « l’Indice de Référence »). Le Fonds a sous-performé son Indice  de Référence au cours du semestre clos le 30 juin 2022. \\

\underline{\textbf{Spanish text}} \\
A título meramente indicativo, la rentabilidad del Fondo se  compara con la del índice Bloomberg Barclays U.S. Government/Credit Bond Index (Total Return) (el «\textbf{Índice de referencia}»). El Fondo tuvo un rendimiento  inferior al de su \textbf{Índice de referencia} en el semestre cerrado a 30 de junio de 2022. \\

\underline{\textbf{Observation:}} the term ``Índice de referencia'' is chosen as a replacement of the full bond name it must be used consistently throughout the rest of the document. \\ 
\hline
\end{tabularx}
\caption{\label{table-examples}
Examples of context-sensitive phenomena in the test set. The words in bold are context-dependent and need extra-sentential information to be correctly translated.
}
\end{table*}

\section{Experiments}
\label{sec:experiment}

\subsection{Experimental setting}

In order to demonstrate the relevance and usefulness of our proposed test set, we use it to evaluate a selection of models, namely the LLMs Llama-3-70b, Llama-3.1-8b, GPT-4o and GPT-4o-mini. 
We do not compare the performance to traditional MT systems because of the limited context size. Most open-source models support a context size of maximum 512 tokens, which would exclude a large proportion of our segments. And as for commercial MT systems such as Google Translate API or DeepL API, the implementation details of whether the systems encode the inputs as a whole are unknown.
That is why we chose LLMs. They have larger context-size limits and we can control whether we want the model to treat the segments as a whole or sentence by sentence. We selected these models in pairs (taking a big and a small one from the same family of models) and we try to answer the following research questions : 
\\
1) \textit{Can the test set show the differences in how models deal with  context?} 
\\
2) \textit{Can the test set show if the models are able to adequately translate specialised financial texts?}

We considered English as the source language for all language pairs, except for French-Spanish where French is considered the source. We translated the test set in two contrastive ways: (1) translating each sentence of a segment one by one, called \textit{``Per sent''} setting and (2) translating each segment as a whole, called the \textit{``Full seg''} setting, hence imitating a sentence-level and document-level translation respectively. To obtain the translations, we used the APIs Groq\footnote{\url{https://wow.groq.com/} to run \texttt{llama3-70b-8192}}, Together\footnote{\url{https://api.together.xyz/} to run \texttt{meta-llama/Meta-Llama-3.1-8B-Instruct-Turbo}} and OpenAI\footnote{\url{https://platform.openai.com/} to run \texttt{GPT-4o} and \texttt{GPT-4o-mini}}. We used a simple prompt to obtain the translations: ``\texttt{Translate the following text in \{src\_lang\} into \{trg\_lang\}. Only provide the translation without any other text. The text to translate:\textbackslash n\{segment\}}''.

We evaluated the translations using Comet-da-22\footnote{Unbabel/wmt22-comet-da} (model that uses the reference as well), with SLIDE approach and a stride of one and window size of three. Since the test set doesn't respect a perfect sentence-to-sentence alignment, we had to approximate the alignment by merging some sentences, in order to be able to use the sliding method.

\subsection{Results}

 \begin{table*}
 \centering
    \begin{tabular}{c|ccc:ccc}      
     \hline 
        Language & Per sent & Full seg & diff & Per sent & Full seg & diff  \\ 
         \hline 
         & Llama-3.1-8b &  &  & Llama-3-70b &  &   \\ 
         \hdashline
        En-Es & 74.14 & 75.22 & +1.08 & 77.71 & 81.51 & +3.79 \\ 
        Fr-Es & 74.48 & 74.90 & +0.42 & 76.46 & 82.03 & +5.57 \\ 
        En-Fr & 74.00 & 73.55 & -0.45 & 77.96 & 82.08 & +4.12 \\ 
        En-It & 74.16 & 73.77 & -0.39 & 78.75 & 82.93 & +4.18 \\ 
        En-De & 69.57 & 61.52 & -8.05 & 79.75 & 81.75 & +2.00 \\ 
        
        \hline 
         & GPT-4o-mini &  &  & GPT-4o &  &   \\ 
         \hdashline
        En-Es & 79.28 & 82.55 & +3.27 & 79.61 & 83.02 & +3.41 \\ 
        Fr-Es & 79.22 & 82.82 & +3.61 & 79.27 & 82.92 & +3.65 \\ 
        En-Fr & 80.72 & 82.87 & +2.15 & 81.01 & 83.37 & +2.36 \\ 
        En-It & 81.57 & 84.54 & +2.97 & 81.86 & 84.87 & +3.01 \\ 
        En-De & 81.39 & 83.75 & +2.36 & 81.78 & 84.24 & +2.46 \\ 
        
        \end{tabular}
    \caption{\label{table:results}Comet-slide scores on the DOLFIN test set.\\
        }
\end{table*}

\textbf{Context-sensitive aspect.} The results of translations of DOLFIN using the Comet-slide metric are shown in Table~\ref{table:results}. 
Since this test set is intrinsically document-level, we hypothesise that the models will perform best when given the full segment, if capable of dealing with long contexts.

We can see that among the four tested models, Llama-3.1-8b is the only one that doesn't improve with full context and even degrades for three language pairs, while its bigger counterpart, Llama-3-70b consistently improves. When analysing the translations of the small model, we observe that for some segments, the generation enters a downhill and with every token the model's predictions get worse and the translation ends in gibberish. This might be explained by the fact that although the model's maximum context size is 128k tokens, this model was fine-tuned from a model with a smaller maximum context size in order to extend it and this didn't give it enough capacity to model long context.

On the other hand, the other bigger models consistently improve when provided with more context. Even in the \textit{Per sent} setting, these models achieve higher scores, which shows their overall higher quality. This gets strengthened when provided with the full context, which shows that they can effectively model context and take advantage of it when generating the full translation. 

The fact that the scores do indeed get higher with more context proves that this test set is appropriate when evaluating document-level MT since it contains context-sensitive phenomena that need additional context to be correctly translated and to obtain higher automatic score.

Apart from the automatic score, the test set allows qualitative manual analysis. To illustrate such an analysis, we performed a simple evaluation (with one evaluator), focusing on segments containing a CTXPRO annotation. This showed many cases where additional context guided the model to generate a more appropriate translation.  As an example, the improvement in translation can be observed in the following sentence: ``These key issues are defined by sector and are regularly reviewed. They are, however, by definition not exhaustive.'' The ``key issues'' were translated in French by Llama-3-70b as ``questions clés''  (feminine noun) and therefore must be referred to by the pronoun ``elles''. The \emph{Per sent} translation contained the masculine ``ils'', which was corrected when provided with the full context. 
Another example of an issue in the \emph{Per sent} setting is the inconsistent level of formality when translating the pronoun ``you''. Within the same section, the model GPT-4o-mini translated it into Spanish by a mix of the informal pronoun ``tú'' and the formal ``usted'', which was also solved in the \textit{Full seg} setting.

To verify the general quality of the test set, we compared the scores obtained on DOLFIN to those on NTREX, a high-quality document-level general domain test set, and they are of similar magnitudes. Due to budget constraints, we computed the translations only with the Llama-3-70b model and we obtained 77.64, 75.93, 77.26 for En-Fr, En-De and En-Es respectively in the \textit{Per sent} setting, which is comparable to the scores on DOLFIN. Since this evaluation method leverages the reference, this shows that the alignment of our sections is correct. If the reference weren't aligned to the source, the scores would be much lower because the translation wouldn't be in line with the reference. We show this by running an evaluation with Comet-slide and using always the same reference to make sure the scores are not high just because the metric relies on the source text. Under these conditions indeed the scores dropped to approximately 44, supporting our hypothesis. We are aware that scores obtained across different test sets are not directly comparable, but we provide this as an additional guarantee of good alignments.

\textbf{Financial aspect.} The test set also allows to evaluate the ability of models to translate financial texts with the required degree of specialisation.
Indeed, the tested models do have issues related to the specialised aspect of the documents. Even when including the full context, some polysemous words are translated incorrectly by the Llama-3-70b model. The word ``Charges'' was translated as ``criminal charges'' instead of ``price'' which is the word's most common usage in finance, giving ``Anklagen'' and ``Anschuldigungen'' in German.
The tested models fail to adhere to a correct formatting style for currency values in the target language and they don't maintain the same format when generating the sentences separately. The translation usually keeps the format as in the source language, English, which is incorrect in most other European languages (like using ``.'' as a decimal separator and placing the ISO currency code before the number).

In terms of table formatting, we find that some models struggle with the markdown syntax. A common problem is a misinterpretation of the horizontal lines (denoted by ``\texttt{---}''), which are misplaced in the translation, resulting in an unreadable table. The content of the table itself is not affected, however in terms of post-editing time and effort, this type of problem degrades the quality of the translation.

\section{Discussion}
\label{sec:discussion}

\textbf{Document-level MT evaluation metrics}

The SLIDE technique was proposed to overcome the limit of maximum context-length of modern evaluation metrics. However, this solution is far from ideal, since it expects a perfect sentence-to-sentence alignment, which doesn't necessarily occur in translations. The evaluation beyond the sentence level remains thus an open issue.

\textbf{Preference of Comet for shorter segments}

When using \emph{Comet-slide} to score the segments, there is a clear tendency of the scores getting lower with increased sentence length. We calculated the correlation between the score and the source/target length and it is of -0.281/-0.283 respectively, which shows a weak negative correlation. This means that as segments get longer, the scores tend to get lower. This must be taken into consideration when using this metric to filter datasets as such a preference for short segments might unnaturally skew the data by stripping the dataset of long sequences.

\textbf{Rich visual formatting in MT}

The formatting remains unsolved in MT, where research mainly focuses on plain text. In this work, we processed PDF files to obtain MD files. While this helps to preserve some information expressed through formatting, it also introduces noise. The final sections can contain formatting unpleasant to the eye and can be made only of a sequence of numbers or a few words, probably extracted from charts and graphs, which makes any linguistic analysis difficult.
The PDF extraction step is a bottleneck. In the case of financial documents with very rich formatting and multi-modal content (charts, graphs, info-graphics, etc.), one would ideally make use of the visual aspects when translating; a multi-modal approach combining vision and language would be interesting to test.

\textbf{Low-resource languages}

We experimented with using Fundinfo to obtain data in a low-resource language, namely Slovak. There were only 73 pairs of documents available, all issued from the same company, which made the texts repetitive and the result was only 8 unique parallel segments. This is the case for most low-resource languages, since only five languages (English, French, German, Spanish, Italian) make up 83\

\section{Future work}

In the future, we aim to provide further manual annotations of the test set with regard to the context-sensitive phenomena to offer the possibility of a targeted automatic evaluation of these aspects.
We will also translate the test set using different MT models (context-aware and context-agnostic) and perform a human evaluation. This evaluation will be used to meta-evaluate existing metrics, to see if 1) they are capable of penalising context-related errors and 2) correctly assess translations in the financial domain. 
The human judgements thus obtained can also be used for training a new metric that would be specialised for financial texts and capable of correctly assessing the models' capability to take context into account.

\section{Conclusions}

We presented DOLFIN, a novel test set for document-level evaluation of MT in the financial domain. It allows to assess the models' capability to translate longer texts while taking into account the context within the financial domain. We described the pipeline created to build it and we release it publicly. In order to illustrate the usefulness of this resource, we used this test set to evaluate a series of models and found that although some of the bigger models benefit from longer contexts, the context might negatively affect the overall quality when the model can't handle it correctly.

\section*{Limitations}
One limitation of our work is that we lack information on the original language of a document. We translate the segments in a ``English-to-any'' fashion (except for French-Spanish), which might cause some of the segments to be less challenging for an MT system if the English text is in fact a translation itself \cite{federmann-etal-2022-ntrex}.

Another identified limitation is that we used the Llama-3 model in order to select the most challenging segments. As any language model, Llama-3 is prone to error, therefore this way of selecting segments is subject to imperfections coming from the model. However, all the segments were also manually verified. Another limitation associated with the use of LLMs is the uncertainty as to whether the texts in the test set have been seen by the models. But the data is present on the web in its PDF format and not in its text and aligned form. Moreover, our study compares the model's performance in the sentence and document-level scenarios, so both scenarios would have the same advantage and therefore the gains or drops in translation quality are still meaningful.

\bibliographystyle{acl_natbib}

\clearpage

\appendix

\section{Prompts used}
\label{annex:prompts}

The English prompt used for identifying elements in English texts:

\noindent\fbox{%
    \parbox{\textwidth}{%
Your task is to analyze a text from a financial document to identify any linguistic phenomena that would make it difficult to translate the text accurately without surrounding context. Imagine the text is being translated sentence-by-sentence in a random order by a professional translator who cannot see the preceding or following sentences. \\
\\
The key types of context-dependent linguistic phenomena to look for are:\\
\\
- Anaphoras: Pronouns that refer back to an antecedent mentioned earlier \\
- Terminology consistency: Terms that need to be translated consistently even if the target language has synonyms \\
- Ellipsis: Omitted parts of the text that refer to something mentioned previously \\
- Polysemous words: Words with multiple meanings where the correct meaning depends on context \\
- Any other phenomena that would make translation ambiguous without more context \\
\\
Here is the text to analyze:\\
\\
<text>\\
{text}\\
</text>\\
\\
Please read the text carefully and identify any of the linguistic phenomena described above. For each phenomenon you find:\\
\\
1. Describe what type of phenomenon it is. \\ 
2. Explain why this particular instance would be difficult to translate accurately without the surrounding context. \\
3. Explain what information would the translator need to determine the correct translation. \\

After analyzing the full text, please provide an overall score from 1-5 indicating how prevalent these context-dependent translation challenges are in the text: \\
 \\
1 = No challenging phenomena identified  \\
2 = One or two minor instances \\ 
3 = Multiple instances that could lead to some ambiguity \\
4 = Significant amount of ambiguity that would make translation difficult \\
5 = Pervasive issues that would make coherent translation nearly impossible without context \\
\\
Provide your 1-5 score surrounded by tags <score> score here </score> without any further explanation.
    }%
}

\clearpage

The French prompt:

\noindent\fbox{%
    \parbox{\textwidth}{%
Ta tâche consiste à analyser un texte tiré d'un document financier afin d'identifier tout phénomène linguistique qui rendrait difficile la traduction du texte sans le contexte. Imagine que le texte est traduit phrase par phrase dans un ordre aléatoire par un traducteur professionnel qui ne peut pas voir les phrases précédentes ou suivantes. \\

Les principaux types de phénomènes linguistiques dépendant du contexte à rechercher sont les suivants :\\
\\
- Les anaphores : Les pronoms qui renvoient à un antécédent mentionné plus tôt. \\
- Cohérence terminologique : Termes qui doivent être traduits de manière cohérente, même si la langue cible contient des synonymes.\\
- Ellipse : Parties du texte omises qui renvoient à un élément mentionné précédemment.\\
- Mots polysémiques : Mots à sens multiples dont le sens correct dépend du contexte.\\
- Tout autre phénomène qui rendrait la traduction ambiguë en l'absence de contexte.\\
\\
Voici le texte à analyser :\\
\\
<texte>\\
{text}\\
</text>\\
\\
Lis attentivement le texte et identifie les phénomènes linguistiques décrits ci-dessus. Pour chaque phénomène identifié :\\

1. Décris de quel type de phénomène il s'agit.\\
2. Explique pourquoi cet exemple particulier serait difficile à traduire avec précision sans le contexte environnant.\\
3. Explique quelles sont les informations dont le traducteur aurait besoin pour déterminer la traduction correcte.\\

Après avoir analysé le texte complet, attribue une note globale de 1 à 5 indiquant le degré de présence de ces difficultés de traduction liées au contexte :\\
 \\
1 = Aucun phénomène problématique identifié \\
2 = Un ou deux cas mineurs\\
3 = Plusieurs cas susceptibles d'entraîner une certaine ambiguïté\\
4 = Beaucoup d'ambiguïté qui rendrait la traduction difficile\\
5 = Problèmes omniprésents qui rendraient une traduction cohérente presque impossible sans contexte\\
 \\
Indique la note de 1 à 5 entourée des balises <score> score ici </score> sans autre explication.
    }%
}

\clearpage

\section{Sub-domains of finance}
\label{annex:sub-domains}

The types of documents produced in the financial sector refer to different sub-domains of finance. Based on the document type, we grouped them into in 4 larger sub-domains: 1)~Fund Prospectus, 2)~KIID - PRRIPS, 3)~Investment Comments and 4)~Fund Annual Report. They are different in their content, style, degree of formality and even terminology. Therefore, sometimes it is advisable to treat the different sub-domains separately. We provide the sub-domain as meta-data for every segment to allow further analysis. The test set contains 3591, 3667, 1344, 1157 segments for each sub-domain respectively. Table \ref{table:stats_subdomains} states the number of segments  per document type.

\begin{table*}[t]
    \centering
    \begin{tabular}{lcc}
        \hline
        \textbf{Document type} & \textbf{\# Seg.} & \textbf{\# Avg Sent.} \\
        \hline
        Prospectus & 3378 & 10.21 \\
        PRIIP Key Information Document & 1974 & 14.13 \\
        Key Information Document & 1693 & 16.56 \\
        Monthly Report & 729 & 12.99 \\
        Annual Report & 719 & 13.21 \\
        Semi-annual Report & 438 & 16.5 \\
        Additional Information for  Investors & 218 & 8.18 \\
        Terms of Contract & 173 & 9.65 \\
        Monthly Manager Commentary & 169 & 15.04 \\
        Fund Profile & 94 & 11.97 \\
        SFDR Pre-Contractual Disclosure Document & 46 & 12.53 \\
        Legal Message & 40 & 8.64 \\
        Supplement Prospectus & 40 & 9.35 \\
        Quarterly Report & 34 & 14.46 \\
        ESG Factsheet & 14 & 11 \\
        \hline
    \end{tabular}
\caption[Caption for LOF]{Statistics per document types, summed among languages. \textit{\# Seg.} denotes the amount of segments within the subdomain, \textit{\# Avg Sent.} the average number of sentences per segment.}\label{table:stats_subdomains}
    \label{table:stats_subdomains}
\end{table*}

\section{Choice of a PDF extractor}
\label{annex:pdf_extractor}

When choosing the PDF extraction tool, the performance of two tools was compared: PyMuPDF and Apryse. This evaluation was done as part of a separate project, the tools were tested in a study for a different task (identification of relevant texts to a query) that also required PDF to text extraction. The evaluation was done manually on 58 documents and also automatically, by measuring how the quality of the texts obtained from the two tools affected the score of the final task. We discovered that texts obtained by PyMuPDF yielded an accuracy score of 51, while texts from Apryse attained 68. The manual analysis showed that this was mainly due to the fact that Apryse is able to detect which text is a title and therefore enables to construct sections, while PyMuPDF only extracts plain text, no matter the functionality of the text within the document. For building our testset we needed the information about the titles and sections, which is why we chose the tool Apryse.

\end{document}